\documentclass[10pt,twocolumn,letterpaper]{article}

\usepackage{cvpr}

\usepackage{amsmath,amssymb,amsfonts}
\usepackage{graphicx}
\usepackage{booktabs}
\usepackage{multirow}
\usepackage{xcolor}
\usepackage{enumitem}
\usepackage{caption}
\usepackage{array}
\usepackage{tabularx}
\usepackage{microtype}

\definecolor{linkblue}{rgb}{0.21,0.49,0.74}
\usepackage[pagebackref,breaklinks,colorlinks,allcolors=linkblue,
  pdftitle={GDP.pdf: Benchmarking Grounded Multimodal Reasoning over Professional PDF Documents},
  pdfauthor={Suhaas Garre, Emily Ritchie, Sushant Mehta, Edwin Chen}]{hyperref}

\def\confName{CVPR}
\def\confYear{2026}

\newcommand{\benchmark}{GDP.pdf}
\newcommand{\blfootnote}[1]{\begingroup\renewcommand\thefootnote{}\footnote{#1}\addtocounter{footnote}{-1}\endgroup}
\newcolumntype{L}[1]{>{\raggedright\arraybackslash}p{#1}}
\newcolumntype{Y}{>{\raggedright\arraybackslash}X}

\title{\benchmark: Benchmarking Grounded Multimodal Reasoning over Professional PDF Documents}

\author{Suhaas Garre\thanks{Correspondence to: \texttt{suhaas@surgehq.ai}}, Emily Ritchie, Sushant Mehta, Edwin Chen  \\   
Surge AI \\
}

\def\pid{}

\begin{document}
\maketitle
\blfootnote{Accepted at the 2nd Workshop on Knowledge-Intensive Multimodal Reasoning (KnowledgeMR) at CVPR 2026.}

\begin{abstract}
A large share of day-to-day work in professional domains happens inside PDF files: benefits packets, leases, datasheets, clinical guidelines, construction plans. Benchmarks for document AI have generally measured the required capabilities in isolation: OCR, layout analysis, chart reasoning, table QA, document VQA. A high score on any one of them does not necessarily reveal whether a model can answer a realistic question that someone in the field would actually ask about a specific PDF. \benchmark{} is a benchmark built to measure this directly. It consists of question--document pairs authored by working professionals in ten fields, and a candidate question was kept only when at least two frontier multimodal models failed it in a way that mattered: a wrong answer, missed decisive evidence, or a fabricated claim, rather than a superficial difference such as style. Each item comes with a rubric of atomic criteria, so we can report a graded rubric score as well as a strict task-level pass rate, and each item is tagged against a taxonomy of eleven capabilities in three tiers, spanning text extraction and grounding, table and chart comprehension, cross-referencing, spatial reasoning, and abstention on unsupported queries. We report results for seventeen frontier models on the 100-item benchmark, current as of July 2026: the best model passes only 30.7\% of the items and the worst passes 2\%. Most errors trace back to a small set of recurring loss patterns: misaligned tables, misread charts, skipped footnotes and exclusions, miscounted floor-plan symbols, scan noise, and amendments that supersede earlier text. The full 100-item benchmark is publicly available at \url{https://huggingface.co/datasets/surgeai/GDP.pdf}.
\end{abstract}

\section{Introduction}

Multimodal models are usually evaluated on visual QA of a fairly academic kind. The document tasks people would actually like to hand to a model look different. Comparing plan tiers in a benefits packet, locating the indemnification clause in a lease, or counting fixtures on a floor plan all require working through a long and variably formatted file, and the fact that settles the answer may sit in a footnote three pages away from the flowchart it qualifies. Current models score well on the standard visual-reasoning suites, but we find those scores to be a poor guide to performance on the document workflows that power everyday economic activity.

Several separate problems are involved here, and prior benchmarks have tended to study each in isolation. The first is document structure (multi-page tables, sidebars, legends, footnotes, amendments appended at the end). The second is background knowledge: a benefits table assumes, for example, that the reader knows what a ``tier'' is. The third problem, which was also a major motivation of this work, is that the failures are not visible as failures. The model cites a clause that exists and a number that is on the page; the clause is simply not the one that governs the user's question.

\benchmark{} was constructed to evaluate all three problems jointly. The questions are phrased as real practitioners phrase them, the input is the original PDF rather than a cleaned-up extract, and the grading checks whether the response rests on the correct evidence. The current version contains 100 items covering ten domains (\emph{Finance}, \emph{Healthcare}, \emph{Legal}, \emph{STEM/Research}, \emph{Engineering}, \emph{Construction}, \emph{Manufacturing/Supply Chain}, \emph{Insurance}, \emph{Real Estate}, and \emph{Human Resources}). Every item has an expert rubric and capability tags, and an item was admitted only after at least two frontier models had failed it. The full 100-item benchmark is released with source PDFs, prompts, rubrics, and domain labels.\footnote{Dataset: \url{https://huggingface.co/datasets/surgeai/GDP.pdf}; evaluation harness: \url{https://github.com/surge-ai/gdp-pdf}.}

\paragraph{Contributions.}
\begin{enumerate}[leftmargin=1.1em,itemsep=0pt,topsep=2pt]
    \item The \benchmark{} benchmark: expert-authored tasks over professional PDFs in ten workflow domains, screened so that every item defeats at least two frontier models.
    \item A taxonomy of eleven capability axes in three tiers (foundational extraction and grounding; structural and multimodal comprehension; advanced reasoning), which is used to tag tasks and to organize the error analysis.
    \item An evaluation protocol built on atomic rubric criteria, reported as a graded rubric score and as a strict pass rate.
    \item An evaluation of seventeen frontier models, with an analysis of the failure patterns we expect will matter most in professional use.
\end{enumerate}

\section{Related Work}

\paragraph{Document parsing, OCR, and layout analysis.}
The early document benchmarks were concerned with turning page images into structure. FUNSD~\cite{jaume2019funsd} covered form parsing, CORD~\cite{park2019cord} covered receipts, PubLayNet~\cite{zhong2019publaynet} and DocLayNet~\cite{pfitzmann2022doclaynet} covered page layout, and DocILE~\cite{simsa2023docile} covered localization and extraction on business documents. Later work has continued in this direction with an emphasis on OCR and layout robustness; see OCRBench~\cite{liu2024ocrbench} and OCRBench v2~\cite{fu2025ocrbenchv2}, OmniDocBench~\cite{ouyang2025omnidocbench}, Real5-OmniDocBench~\cite{zhou2026real5}, RoDLA~\cite{chen2024rodla}, and MMDocBench~\cite{zhu2024mmdocbench}. All of these measure how well a model reads the page. None of them measures whether a system can use what it reads to complete a task someone is paid to do, which is the question we wanted to answer.

\paragraph{Document QA, long documents, and retrieval-aware systems.}
The document VQA line moved evaluation from parsing toward question answering: DocVQA~\cite{mathew2021docvqa}, InfographicVQA~\cite{mathew2022infographicvqa}, MP-DocVQA~\cite{tito2023mpdocvqa}, DUDE~\cite{vanlandeghem2023dude}, and TAT-DQA~\cite{zhu2022tatdqa}. DocBench~\cite{zou2024docbench}, MMLongBench-Doc~\cite{ma2024mmlongbench}, LongDocURL~\cite{deng2025longdocurl}, and M-LongDoc~\cite{chia2024mlongdoc} extended the setting to long documents and raw files, and the retrieval-oriented papers (PDFTriage~\cite{saadfalcon2024pdftriage}, MMDocRAG~\cite{dong2025mmdocrag}) argued that finding the evidence inside a long multimodal file is a large part of the problem in its own right. \benchmark{} overlaps most with this group. The difference lies in what the questions assume of the reader. Ours concern professional documents where the answer hinges on footnotes, exclusions, legends, or superseded sections, and where a reader is expected to know what those constructs mean.

\paragraph{Charts, tables, and domain-specific reasoning.}
Charts have dedicated benchmarks (ChartQA~\cite{masry2022chartqa}, PlotQA~\cite{methani2020plotqa}, CharXiv~\cite{wang2024charxiv}, ChartQAPro~\cite{masry2025chartqapro}), as do tables and numerical reasoning (HybridQA~\cite{chen2020hybridqa}, TAT-QA~\cite{zhu2021tatqa}, FinQA~\cite{chen2021finqa}). FinanceBench~\cite{islam2023financebench}, LegalBench~\cite{guha2023legalbench}, and ClinicBench~\cite{liu2024clinicbench} test financial, legal, and clinical knowledge respectively. What nearly all of these have in common is that the model receives a cleaned input -- extracted text, one table, or one chart by itself. By the time the input has been cleaned to that degree, much of what makes a real PDF difficult has already been removed.

\paragraph{Broad multimodal evaluation.}
The general multimodal suites (MMMU~\cite{yue2024mmmu}, MMMU-Pro~\cite{yue2024mmmupro}, MathVista~\cite{lu2024mathvista}, MEGA-Bench~\cite{chen2024megabench}) track progress across many task types at once, and the knowledge-focused VQA datasets, OK-VQA~\cite{marino2019okvqa} and A-OKVQA~\cite{schwenk2022aokvqa}, test world knowledge over natural images. These benchmarks serve a purpose different from ours. None of them was designed to indicate whether a model can be trusted with a benefits packet or a deed. Table~\ref{tab:positioning} summarizes where \benchmark{} sits relative to these families.

\begin{table*}[t]
\centering
\setlength{\tabcolsep}{4pt}
\begin{tabularx}{\textwidth}{L{2.25cm}L{5.9cm}Y}
\toprule
\textbf{Benchmark family} & \textbf{Representative work} & \textbf{Evaluates well} \\
\midrule
Layout / OCR / parsing & FUNSD, CORD, PubLayNet, DocLayNet, DocILE, OmniDocBench & How well the page is read: OCR quality, layout detection, extraction \\
Document QA / long-doc & DocVQA, MP-DocVQA, DUDE, DocBench, MMLongBench-Doc, M-LongDoc & Question answering over raw or long documents, often with evidence spread over many pages \\
Chart / table / domain & ChartQA, CharXiv, TAT-QA, FinQA, FinanceBench, LegalBench & Charts, tables, or one specialized domain, usually with a cleaned input \\
Broad multimodal & MMMU, MMMU-Pro, MathVista, MEGA-Bench & General multimodal reasoning across many task types \\
\textbf{This work} & \textbf{\benchmark} & \textbf{Whether answers to professional questions are grounded in the right evidence from the original PDF} \\
\bottomrule
\end{tabularx}
\caption{\benchmark{} relative to neighboring benchmark families.}
\label{tab:positioning}
\end{table*}

\section{\benchmark: Benchmark Design}

\subsection{Task Formulation}

Each benchmark item is a tuple
\[
x_i = (P_i, q_i, R_i, a_i, d_i, \tau_i),
\]
where $P_i$ is a PDF document, $q_i$ is a natural-language question, $R_i = \{r_{ij}\}_{j=1}^{K_i}$ is an expert rubric containing $K_i$ atomic criteria, $a_i$ is an expert reference answer used during rubric authoring, $d_i$ is a domain label, and $\tau_i \subseteq \mathcal{T}$ is a set of capability-axis tags drawn from the taxonomy $\mathcal{T}$.

Given a model response $\hat{y}_i = M(P_i, q_i)$, the grader assigns a binary score $g_{ij} \in \{0,1\}$ to each rubric element $r_{ij}$. The item-level rubric score is
\[
s_i(M) = \frac{1}{K_i}\sum_{j=1}^{K_i} g_{ij},
\]
and the benchmark-level mean rubric score is
\[
\mathrm{Score}(M) = \frac{1}{N}\sum_{i=1}^{N} s_i(M).
\]
We also define a strict pass-rate indicator
\[
p_i(M)=
\begin{cases}
1 & \text{if } g_{ij}=1 \text{ for all } j\\
0 & \text{otherwise,}
\end{cases}
\]
with benchmark-level strict pass rate
\[
\mathrm{PassRate}(M)=\frac{1}{N}\sum_{i=1}^{N} p_i(M).
\]
Models are ranked by strict pass rate; the mean rubric score supports finer-grained analysis. An item passes strictly only if every element of its rubric is satisfied; Table~\ref{tab:mainresults} reflects how rarely that happens. The rubric score keeps the partial credit and shows the sub-tasks and criteria that the model gets right.

\subsection{Design Principles}

\noindent\textbf{Prior knowledge.}
If general prior knowledge suffices to answer a question, the task was rejected: the answer has to depend on the attached PDF.

\noindent\textbf{Realism.}
The questions came from contributors' actual jobs and workflows. Each contributor was encouraged to submit the kind of question they would actually have typed to an assistant mid-task at work.

\noindent\textbf{Adversarial construction.}
Candidate tasks were attempted by several frontier multimodal models, and only candidates on which at least two made a major, meaningful error were included. Because screening draws on a pool of models rather than a fixed pair, item selection is not tied to the failure profile of any single model or family.

\noindent\textbf{Knowledge-intensive grounding.}
Tasks in which perception and domain knowledge interact were favored. Professional documents can put a surprising amount of weight on fine print: exclusions, footnotes, legends, plan symbols, amendments.

\noindent\textbf{Diagnostic granularity.}
Capability tags were added to every item, so the results can be sliced by failure type.

\subsection{Benchmark Profile and Coverage}

Table~\ref{tab:profile} summarizes the details of the current version of \benchmark{}. We considered heterogeneity in both domain and in artifact type, so \benchmark{} includes financial filings and benefits packets, datasheets and clinical guidelines, floor plans, insurance policies, deeds, inspection reports and scanned amendments.

\begin{table}[t]
\centering
\small
\setlength{\tabcolsep}{3pt}
\begin{tabularx}{\columnwidth}{L{2.65cm}Y}
\toprule
\textbf{Property} & \textbf{Details} \\
\midrule
Domains & 10 professional domains, evenly distributed \\
Prompt curation & Questions written by domain experts from their own work, with the original PDFs and the collectors' notes \\

Document types & Policies and forms; multi-page tables; charts and infographics; datasheets; floor plans; long scanned deeds and guidelines \\

Screening rule & An item stays only if at least two frontier models commit a major failure on it \\

Annotations & Reference answer, atomic rubric, domain label, capability tags, notes on the parsing challenge \\

Question styles & Lookup, comparison, reconciliation, calculation, summarization; some queries are deliberately unsupported \\

Reporting metrics & Mean rubric score for analysis; strict pass rate for comparing models \\

Public release & All 100 examples on Hugging Face, released under Apache~2.0; third-party PDFs retain their original rights \\
\bottomrule
\end{tabularx}
\caption{Profile of the current version of \benchmark{}.}
\label{tab:profile}
\end{table}

Table~\ref{tab:domainprofile} breaks the coverage down by domain. As previously mentioned, any task that was answerable without reading the document was rejected.

\begin{table*}[t]
\centering
\scriptsize
\setlength{\tabcolsep}{3.5pt}
\begin{tabularx}{\textwidth}{L{1.55cm}L{4.0cm}Y}
\toprule
\textbf{Domain} & \textbf{Representative PDFs} & \textbf{How models typically fail} \\
\midrule
Finance & Earnings releases, investor filings, analyst materials & Values come from the adjacent column or the wrong note field; a table which crosses pages tends to lose its alignment. \\
Healthcare & Reviews, dosage tables, clinical guidelines & The model loses its place in a long review, skips a figure footnote, or answers when the information is simply not there. \\
Legal & Contracts, leases, policy language, filings & The cited clause is usually real. It is just not the governing one, and cross-references between sections get dropped. \\
STEM / Research & Climate reports, technical reports, scientific figures & Misread legends, wrong values off the charts, and evidence from several figures that never gets combined. \\
Engineering & Datasheets and specification sheets & Log-scale plots defeat the value reading; units get mixed up; merged headers scramble the lookup. \\
Construction & Floor plans and schedules & Symbols never get matched to the legend, so fixture and window counts come out wrong, and the schedule is never reconciled with the drawing. \\
Manufacturing & Process notes, packing lists, certificates of conformance, technical instructions & Footnotes get skipped, and where the instructions disagree with the model's priors, the priors usually win. \\
Insurance & Auto policies and endorsements & Reading stops at the insuring agreement, short of the exclusions; defined terms get taken at their everyday meaning. \\
Real Estate & Valuation reports, inspection reports, deeds, amendments & Status columns are misread, and superseded sections get quoted as if they were still in force. \\
HR & Benefits packets, leave-policy tables, personnel handbooks & The wrong tenure band gets read, and a date hidden in a footnote silently breaks the chronology. \\
\bottomrule
\end{tabularx}
\caption{Coverage by domain in \benchmark{}.}
\label{tab:domainprofile}
\end{table*}

\subsection{Capability Taxonomy}
\label{sec:taxonomy}

Table~\ref{tab:taxonomy} lists the eleven capability axes, organized into three tiers (extraction and grounding; structural and multimodal comprehension; advanced reasoning). We solicited prompts against this taxonomy, and the error analysis in Section~\ref{sec:failure} is written in its vocabulary. Most items carry more than one tag.

\begin{table*}[t]
\centering
\small
\setlength{\tabcolsep}{4pt}
\begin{tabularx}{\textwidth}{L{2.75cm}L{3.95cm}Y}
\toprule
\textbf{Family} & \textbf{Axis} & \textbf{What the axis measures} \\
\midrule
\multirow{3}{*}{\parbox{2.75cm}{Tier 1: Extraction \& grounding}}
& Correctness \& completeness & The text the answer needs is extracted in full, with no dropped content and no altered facts \\

& Grounding & Every asserted fact can be checked against the PDF; nothing is supplied from prior knowledge or hallucinated \\

& Spatial awareness & When the prompt depends on location, the model knows where pieces of content lie on the page relative to one another \\

\midrule
\multirow{4}{*}{\parbox{2.75cm}{Tier 2: Comprehension (structural \&\\multimodal)}}
& Semantic reading flow & Columns, sidebars, and callouts are read in the order a person would read them \\
& Typographic hierarchy & Headings, bullets, emphasis, and section boundaries mean what they should \\
& Standard table parsing & Rows and columns stay aligned, merged headers and adjacent cells included \\
& Chart \& multimodal interpretation & Legends, axes, and visual marks are interpreted correctly in context \\
\midrule
\multirow{4}{*}{\parbox{2.75cm}{Tier 3: Advanced\\reasoning}}
& Complex \& multi-page tables & Context carries over page breaks, nested headers, and merged cells \\
& Cross-referencing & Footnotes, citations, appendices, and distant definitions are connected to the text they modify \\
& Artifact \& noise & Scan noise, running headers and footers, watermarks, and superseded content are correctly identified as such \\
& Unsupported queries & Where information is missing, redacted, or unsupported, the model states this instead of answering incorrectly \\
\bottomrule
\end{tabularx}
\caption{The capability taxonomy used to author, tag, and analyze \benchmark{} tasks.}
\label{tab:taxonomy}
\end{table*}

\subsection{Collection and Curation Workflow}

\begin{enumerate}[leftmargin=1.3em,itemsep=1pt,topsep=2pt]
    \item \textbf{Candidate sourcing.} A domain expert submits a task from their own work: the source PDF, a note on why the task matters in that line of work, and a first-pass reference answer.
    \item \textbf{Challenge calibration.} Several frontier models (the screening models) attempt the candidate, and at least two of them have to commit a \emph{major failure} for it to advance. Major means a materially wrong final answer, a dropped piece of decisive evidence, or a fabricated claim. Superficial differences in phrasing do not count as a failure.
    \item \textbf{Novelty and ambiguity filtering.} A candidate is rejected at this stage if it is answerable from general priors, rests on context only the original contributor had, or is worded so ambiguously that the semantics of the question, rather than the document, becomes the bottleneck.
    \item \textbf{Rubric authoring.} The expert writes atomic criteria covering the facts the answer must contain, the formulations that count as equivalent, and the claims the answer must not make. For abstention items the rubric states what the abstention has to say.
    \item \textbf{Axis tagging and worker notes.} Capability tags are added with a note describing the parsing challenge in plain terms: merged cells, a footnote, a superseded section, a legend.
    \item \textbf{Sanity checks.} The gold answer must pass its own rubric; a deliberately bad output must fail on the criterion the item was built around.
\end{enumerate}

\subsection{Data Curation Insights}

Assembling the benchmark produced several observations worth recording. Length was a poor predictor of difficulty; several of the hardest items are one-line questions about a single table or figure. Comparisons across two sections, and amendments overriding an earlier line, produced confident hallucination in the best models far more often than expected. Table~\ref{tab:examples} gives a sample of items, the evidence deciding each, and the usual way most frontier models went wrong.

\begin{table*}[t]
\centering
\scriptsize
\setlength{\tabcolsep}{3.5pt}
\begin{tabularx}{\textwidth}{L{1.65cm}L{3.9cm}L{4.2cm}Y}
\toprule
\textbf{Domain} & \textbf{Task sketch} & \textbf{Document evidence} & \textbf{Typical failure mode} \\
\midrule
HR & Order the first five companies to offer paid bereavement leave. & A multi-page table must be sorted by announcement date, but a footnote changes Mastercard's date and removes it from the top five. & The footnote is ignored, so the list comes back fluent but chronologically wrong. \\
Manufacturing & Recommend drill-profile features for cutting holes in Ti-8Al-1Mo-1V pipe. & The relevant section advises against brad-point bits and recommends W+R thinning with a $180^\circ$ chamfer angle. & General machining priors override the document; models recommend brad-point bits and volunteer process advice that was not requested. \\
Healthcare & Identify the EDS subtype(s) associated with \textit{TNXB} mutations and distinguish them. & The review supports clEDS, states that hEDS lacks a diagnostic genetic marker, and does not provide enough evidence for the requested hEDS comparison. & \textit{TNXB} gets presented as if it were a marker for hEDS, and the key clinical distinction is reversed or dropped. \\
Insurance & Decide whether Medical Payments and UMBI apply after a highway collision involving a full-time RV. & Exclusions remove a vehicle used as a residence from the ``uninsured motor vehicle'' definition and bar medical-payments coverage for injuries it causes. & Reading stops at the insuring agreement; the exclusions are missed and coverage is granted incorrectly. \\
Engineering & Read threshold irradiance for a specified ambient brightness and wavelength from a datasheet. & The answer requires reading a precise point from a plotted curve with logarithmic scales and the correct source wavelength. & The right figure is identified, but the values read off the plot are implausible. \\
Construction & Determine how many of the most prominent window type appear on the first floor. & The schedule must be read correctly to identify the dominant type, then matched to plan labels on the first-floor drawing. & A wrong schedule row or quantity column is chosen, and the error propagates into the plan count. \\
Real Estate & Decide whether three ``PENDING'' comparable sales would change the automated valuation. & ``PENDING'' appears in the registration-date column; the sale prices are already reflected, and the PDF does not justify recalculating the estimate. & Pending registration is treated as missing price evidence, leading to a predicted value change the document does not support. \\
\bottomrule
\end{tabularx}
\caption{Representative items. In most cases the decisive evidence is small and local, and is easy to miss even when the overall topic of the document is understood.}
\label{tab:examples}
\end{table*}

\section{Evaluation Protocol}

\paragraph{Atomic rubric design.}
\benchmark{} deliberately does not adopt ANLS-style overlap scoring~\cite{biten2019scene,mathew2021docvqa,peer2024anls}. A response can name nearly the same companies as the reference, in nearly the same words, and still be completely wrong, because the one company a footnote disqualifies is in the list, and an overlap metric will readily reward it. Every item therefore carries a rubric of atomic yes/no criteria.

\paragraph{Unsupported queries and abstention.}
Some tasks ask for what the document does not contain: the figure is redacted, the value never reported, the conclusion does not follow. To get credit the model has to plainly state this instead of hallucinating. Most models instead produce a fluent, helpful-sounding fabrication, which is scored zero.

\paragraph{Slice reporting.}
For any subset of items $\mathcal{I}$ defined by domain, tier, or capability axis, we report
\[
\mathrm{Score}_{\mathcal{I}}(M)=\frac{1}{|\mathcal{I}|}\sum_{i\in\mathcal{I}} s_i(M).
\]
For slices the natural metric is the rubric score, which retains the partial progress open-ended items produce; the failure analysis in Section~\ref{sec:failure} draws on these slices qualitatively. For ranking models against each other we use the strict pass rate (Table~\ref{tab:mainresults}); that number credits fully solved items and nothing else.

\paragraph{Judging and manual verification.}
Rubric criteria are graded by an LLM judge (Gemini 3.5 Flash), calibrated to ensure high agreement with expert human raters. Each criterion is passed to the judge separately and receives a binary pass/fail. Criteria are written to be self-contained, referencing only what should or should not appear in the final response, so the judge is given neither the source PDF nor the gold answer; the gold answer is used while authoring and sanity-checking the rubric but nowhere else, so that it cannot anchor the judgment. The judge's model family also appears among the evaluated models; we note that the criterion-level grading task is far more constrained than the generation task. The grading implementation is released with the benchmark.\footnote{\url{https://github.com/surge-ai/gdp-pdf}} Every failure discussed in Section~\ref{sec:failure} was also checked by hand against the source document and the collector's notes.

\section{Evaluation}
\label{sec:pilot}

\subsection{Evaluated Models}

We evaluate frontier models through their public APIs, with reasoning-effort configurations as listed in Table~\ref{tab:mainresults}. Each model is given the question and the source PDF for all 100 items, with the PDF supplied natively as a base64-encoded file input; each provider's own document handling is therefore part of what is measured. No tools and no additional context are provided. Each item is run five times per model, and we report the strict pass rate averaged over the runs (mean pass@1). The benchmark was released on April 14, 2026, and is re-run as new models ship; Table~\ref{tab:mainresults} reports the leaderboard snapshot as of July 2026, and the qualitative failure analysis in Section~\ref{sec:failure} draws on transcripts from these evaluation rounds.

\subsection{Main Results}

\begin{table}[t]
\centering
\small
\caption{Leaderboard on \benchmark{} as of July 2026. We report strict pass rate: the percentage of items on which a model satisfies \emph{all} required atomic rubric criteria, averaged over five runs per model. No evaluated model passes a third of the items.}
\label{tab:mainresults}
\begin{tabularx}{\columnwidth}{@{}Yc@{}}
\toprule
\textbf{Model} & \textbf{Pass rate (\%)} \\
\midrule
GPT-5.6 Sol & 30.7 \\
Claude Fable 5 (Adaptive Max) & 29.8 \\
GPT-5.5 (xHigh reasoning) & 26 \\
GPT-5.6 Terra & 24.7 \\
Claude Opus 4.8 (Adaptive Max) & 24 \\
GPT-5.6 Luna & 22.7 \\
Claude Opus 4.7 (Adaptive Max) & 21 \\
Claude Sonnet 4.6 (Adaptive Max) & 18 \\
Gemini 3.1 Pro & 17 \\
Gemini 3.5 Flash & 14 \\
Grok 4.5 (High reasoning) & 14 \\
Kimi K2.6 & 12 \\
Gemini 3 Flash & 10 \\
Grok 4.3 (High reasoning) & 8 \\
Mistral Large 3 & 2 \\
Nova 2 Pro & 2 \\
Nemotron 3 Nano Omni & 2 \\
\bottomrule
\end{tabularx}
\end{table}

Table~\ref{tab:mainresults} reports the model scores.\footnote{Table~\ref{tab:mainresults} mirrors the public \benchmark{} leaderboard as of July 2026. The leaderboard at \url{https://surgehq.ai/benchmarks/gdp-pdf} is re-run as new models and inference configurations become available, so the live numbers evolve over time; readers should consult it for the latest results.} No frontier model passes even a third of the items. The best model fails more than two items out of every three, and three models pass just 2\% of the items.

The hardest task slices were the spatial ones (such as floor plans and construction drawings), dense technical plots, and long noisy documents carrying amendments. These show up across the construction, engineering, manufacturing, insurance, and real-estate tasks. On HR and STEM/Research tasks, models scored higher when the task was a lookup in a single table or figure, but a footnote or a cross-page dependency was usually enough to break those as well.

\subsection{Failure Analysis}
\label{sec:failure}

\paragraph{Table alignment failures.}
The most common error was reading the wrong cell, wrong row or wrong column, and it concentrated in tables with merged cells or stacked headers. For example, an HR benefits item asks how plan costs change after an employee adds a dependent. One model picked the wrong tenure band. A second model picked the right band and then reported numbers which are not in it.

\paragraph{Chart and figure misreads.}
The model finds the right figure but misreads it. On an engineering datasheet, the models that failed it during screening located the correct threshold-irradiance plot and returned values that fall well outside the plausible range of the plotted curves.

\paragraph{Footnotes, cross-references, and exclusions get dropped.}
In professional documents the controlling fact often sits in a footnote, or in a definition pages away from where the question gets answered. The bereavement-leave item contains a footnote which moves one company's date and reorders the ranking; most models never used it. In an insurance item, the models identified the at-fault driver correctly, then stopped at the insuring agreement and never reached the exclusions which decided whether the coverage applied.

\paragraph{Prior knowledge overrides the document.}
Where the document and the training prior disagree, the prior tended to win. As an example, every model when asked about drilling Ti-8Al-1Mo-1V pipe recommended brad-point bits. As general machining advice this is reasonable. The document in question, however, advises against brad-point bits for this application and calls for W+R-type thinning with a $180^\circ$ chamfer angle, so the answers sounded helpful and were wrong for the specific document at hand.

\paragraph{Spatial problems are especially challenging.}
A floor-plan question forces the model to match small symbols to a legend and keep track of them from page to page, and the models could rarely do this without errors. On the lighting question they reported fixtures in rooms which have none and missed fixtures which exist. On the window question they miscounted, because the schedule on one page never got connected to the labeled first-floor plan on the next.

\paragraph{Noise and supersession can cause cascading errors.}
Long scanned files with amendments were challenging from the first round of curation to the last. Models quoted deed sections which an amendment had already replaced. In one automated valuation report, every model in our initial evaluation round treated a ``PENDING'' label as though it undermined the associated sale prices. The label sits in the registration-date column and should have had no bearing on the estimate.

\section{Discussion}

\paragraph{Standard scores can be misleading.}
The models in Table~\ref{tab:mainresults} sit at or near the top of most public multimodal leaderboards, yet the best two pass only 30.7\% and 29.8\% of \benchmark{} items. The gap has a simple explanation: \benchmark{} checks something broader suites can skip, namely where the answer came from. One caveat follows from the construction: because every item defeated at least two frontier models at collection time, absolute pass rates measure performance on adversarially selected professional tasks, not on professional document work at large.

\paragraph{The interaction of perception and domain knowledge.}
Most failures could not be cleanly classified into perception errors and reasoning errors. Consider the RV insurance item: the models read the insuring agreement correctly, so perception was fine, and they granted coverage anyway, because they never went looking for the exclusion a policy reader would know to check.

\paragraph{A benchmark for the floor, not just the ceiling.}
Broader benchmarks mostly ask a ceiling question: how hard a task can this model accomplish in a controlled academic setting? Deploying models for document automation requires asking the floor question instead: what can the model be counted on to get right every single time? The 2--30.7\% range in Table~\ref{tab:mainresults} is our answer for the present moment, and it is not a number that supports unsupervised use.

\paragraph{The document work behind economic activity.}
The tasks in \benchmark{} are not niche edge cases. An insurance adjuster reading exclusions, an HR analyst sorting leave announcements, an estimator counting windows: each task appears in this paper because a contributor actually does that work. Current broader benchmarks sample these skills thinly, and we suspect that explains much of the gap between leaderboard capability and practical reliability.

\paragraph{Implications for model development.}
Longer context windows alone might not fix this gap in model capabilities. The failures we saw would respond better to page representations that keep structure intact, to parsers that get charts and tables right, and to retrieval that does not strip away the visual evidence. Footnotes and amendments need to be treated as content rather than noise. And a model with better uncertainty calibration would have scored considerably better on our abstention items.

\section{Conclusion}
We presented \benchmark{}, a benchmark for multimodal reasoning grounded in professional PDF documents, built from expert-authored tasks with the original files and their domain semantics intact. Seventeen frontier models have been evaluated as of July 2026, and none passes a third of the items on the strict metric. The failures concentrated where professional documents are hardest to read: tables, charts, footnotes, exclusions, noisy scans, amendments. We hope the benchmark proves useful for measuring whether models can handle the routine document work that a significant share of professional activity depends on.

{\small
\bibliographystyle{ieeenat_fullname}
\bibliography{gdp_pdf_refs}
}

\end{document}